\documentclass{article}

\usepackage{authblk}
\usepackage{indentfirst}
\usepackage{graphicx}
\usepackage{hyperref}
\usepackage{cite}
\usepackage{subfigure}

\title{Evaluation of Automatic Text Summarization using Synthetic Facts}

\author[1]{Jay Ahn}
\author[2]{Foaad Khosmood}
\affil[1, 2]{Department of Computer Science \& Software Engineering, California Polytechnic State University}
\date{\today}

\begin{document} 
\maketitle
\setlength{\parindent}{20pt}

\section{Introduction}
Despite some recent advances, automatic text summarization remains unreliable, elusive and of limited actual use in applications. Two main problems with current summarization methods are well known: evaluation and factual consistency. First, the current evaluation metrics are inadequate. They rely on human generated summaries which are wildly inconsistent and subjective. Second, there is no way to guarantee generated summaries have true facts consistent with the source text. Some techniques using large language models may introduce new facts into the summary. This latter problem renders automatic summarization  unsuitable in a number of important domains such as law, journalism and government. 

In this paper, we propose a new automatic reference-less text summarization evaluation system that can measure the quality of any text summarization model based on factual consistency, comprehensiveness, and compression rate. The system works by generating dynamic, natural language documents, based on a set of generated facts from a certain domain. These documents are to be used as test input for any automatic text summarization system. Our working hypothesis is that since the input documents contain facts that are entirely known to the system, verifying factual consistency should be more feasible. 

However, factual consistency isn't the only criteria for summarization. To automatically measure the quality of a summary text, we consider three distinct goals of a good summarization system: consistency, comprehensiveness and compression. We define Consistency as the ratio of original source facts and facts appearing in the summary. Comprehensiveness is a score based on how much of the source facts are retained in the summary. Finally, compression is a simple measure of how much shorter the summary text is compared to the original source. While consistency and comprehensiveness scores are to be maximized, the compression score is to be minimized. This provides the basic tradeoff between larger summaries with more factual coverage and smaller summaries with fewer facts reproduced.

As far as we know, we are the first to suggest a testing system that assesses text summarization models with factuality and information coverage.

\section{Background}
Automatic text summarization has achieved remarkable success with the help of deep neural networks and the development of standardized benchmark datasets. It can generate fluent, human-like summaries. However, the unreliability of the existing evaluation metrics/systems has hindered its practical usage and slowed down its progress. The current standard N-gram overlaps-based evaluation metrics such as ROUGE \cite{lin-2004-rouge} or BLEU \cite{papineni-etal-2002-bleu} have shown their limitations in measuring factual consistency \cite{2019, zhou-etal-2021-detecting, zhao2020reducing, goyal-durrett-2020-evaluating, kryscinski-etal-2020-evaluating} and information overlap \cite{deutsch-roth-2021-understanding} between the source document and its output summary. Also, as they rely on human-generated reference summaries, they are not only expensive and noisy but also prone to reference bias \cite{kedzie2019content}. To tackle these issues, researchers have introduced diverse reference-free approaches to evaluate desirable aspects of summarization such as factual consistency and information coverage.

Zhao et al.\cite{zhao2020reducing} introduces entity-based factual consistency measurement by verifying if any entities in the output summary match with those in the source document. Similarly, Xu et al.\cite{xu-etal-2020-fact} suggests a fact-based evaluation by checking if facts in the output summary also appear in the source document. Moreover, QA-based \cite{scialom-etal-2019-answers, wang-etal-2020-asking, durmus-etal-2020-feqa}, sentence-level \cite{kryscinski-etal-2020-evaluating}, or more fine-grained token-level \cite{zhou-etal-2021-detecting} factual consistency evaluation metrics have been proposed with the advancement of natural language understanding using transfer learning. However, these approaches require a hard assumption that the trained neural models demonstrate a solid understanding of the text to trust the extracted facts or their embeddings. Furthermore, these neural-based approaches do not provide the natural interpretability of their evaluation scores which is especially crucial in measuring factual consistency.

When it comes to measuring information coverage, the manual pyramid method \cite{nenkova-passonneau-2004-evaluating} has been served as a gold standard metric. Since manual evaluation entails time and cost, \cite{gao-etal-2018-pyreval, gao-etal-2019-automated} have simply automated the pyramid methods, and \cite{eyal-etal-2019-question, deutsch2021questionanswering} have recently proposed evaluation metrics that highly correlate with the pyramid method, adopting question answering (QA) models. However, these approaches are reference-based; therefore, they are susceptible to reference bias. Hence, \cite{louis-nenkova-2009-automatically} has fitted a linear regression with features such as KL-divergence to measure the information overlap without reference. More recently, \cite{wu-etal-2020-unsupervised, egan2021play} have used pre-trained language models such as BERT to measure the information overlap between the source and its system summary. But, these approaches still carry the aforementioned limitations as neural-based measurements.

To address these issues, we propose an automatic text summarization evaluation metric by utilizing synthetic documents with dynamically generated facts. Since our system generates the source documents with the guarantee to contain only those facts, it is much more reliable than the other approaches that have limitations in extracting all the facts from the complex sentences in the source document. Moreover, it has natural interpretability as it explicitly demonstrates the facts used in the evaluation.

\section{Approach}
The main constraint of the current natural language understanding is its difficulty in understanding and extracting all the facts from a source document which normally contains complex sentences. So, our system instead creates a synthetic source document in different domains by dynamically generating relevant facts with the style and language commonly used in those domains. Then, it measures factual consistency and comprehensiveness by comparing the facts in the summary with those in the source document and produces an overarching quality score with a high compression rate penalty. Our system still requires extracting facts from the summaries, but since they are 100\% known and contain the same factual relations that were used in generating the synthetic source document, they should be easier to detect.

\begin{figure}[htp]
  \centering
  \includegraphics[width=\linewidth]{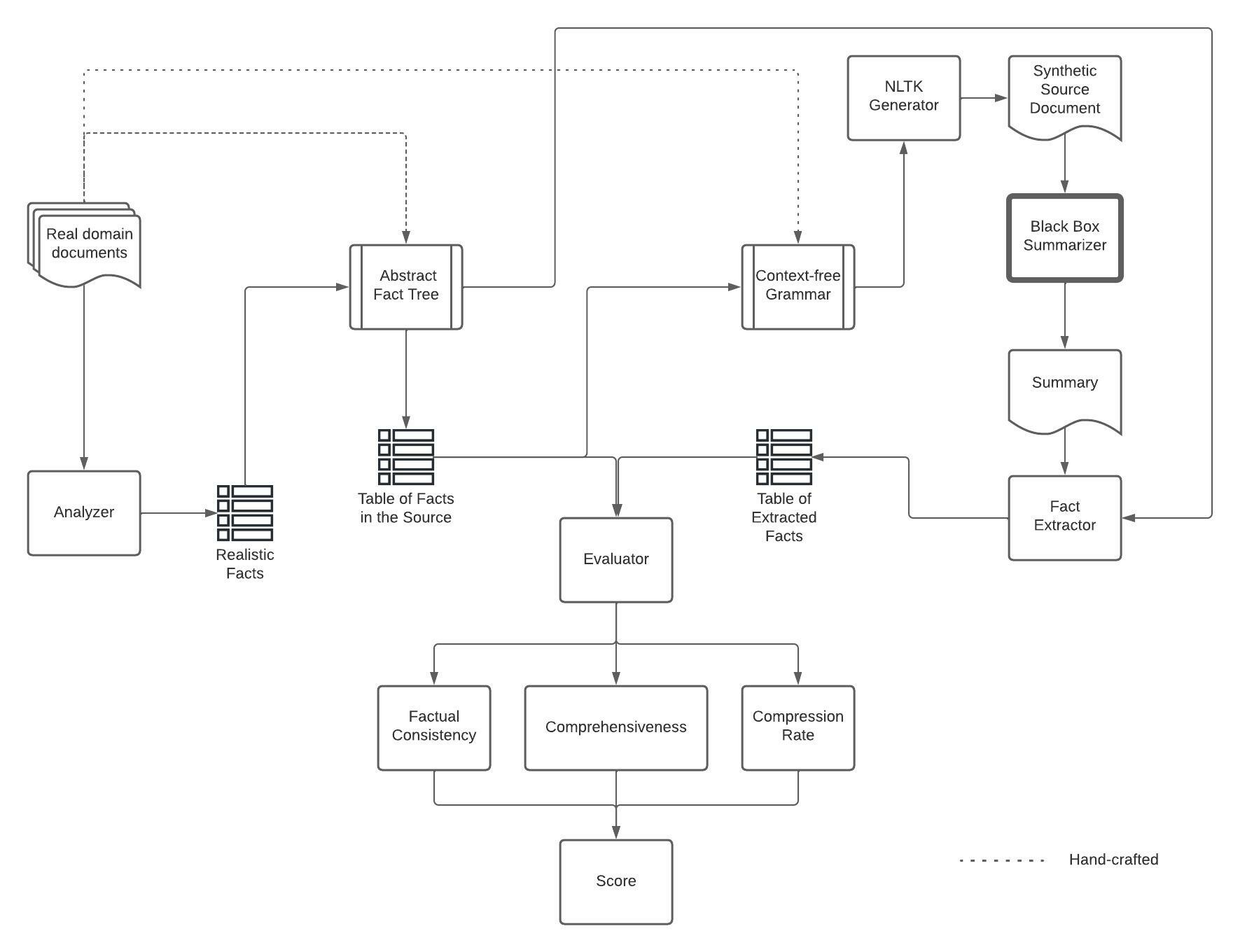}
  \caption{The overview of our text summarization evaluation system}
  \label{fig:sys_design}
\end{figure}

\autoref{fig:sys_design} describes the overview of our system. Based on the real-domain documents, we analyze and mine the most likely facts to generate realistic synthetic documents. Also, we manually create abstract fact trees and context free grammars to write the realistic sentences. Then, with the most probable facts and the predefined abstract fact trees, we populate the table of facts that would appear in the source document and fill the predefined grammars with those facts. Then, we generate synthetic source document using NLTK\cite{bird2009natural} with those grammars. After generating the synthetic document, it gets passed to a black box deep text summarizer to produce the system summary. The fact extractor then extracts the facts from the summary with the same factual relations used in the abstract fact tree and creates a table of the extracted facts. Lastly, the evaluator measures the factual consistency and comprehensiveness with high compression rate penalty by comparing the table of facts in the source and that of those in the summary.

\subsection{Fact Representation}
Facts in a sentence are the realistic lexicons with various linguistic patterns that capture the meaning of the sentence. They are generated with predefined context free grammars, and the total list of the linguistic relations is described in \autoref{tab:fact_table}. A (\textit{main clause, subordinate clause}) relation is simply represented as ((\textit{main noun, main verb}), (\textit{subordinate noun, subordinate verb})).

\begin{table}
\begin{center}
\begin{tabular}{||c||} 
 \hline
 Relation \\
 \hline\hline
 (\textit{subject, verb, object})\\ 
 \hline
 (\textit{noun modifier, noun})  \\
 \hline
 (\textit{verb modifier, verb}) \\
 \hline
 (\textit{phrase modifier, verb}) \\
 \hline
 (\textit{phrase modifier, noun}) \\
 \hline
 (\textit{clause modifier, verb}) \\
 \hline
 (\textit{clause modifier, noun}) \\
 \hline
 (\textit{main clause, subordinate clause}) \\
 \hline
\end{tabular}
\caption{\label{tab:fact_table}A list of linguistic relations used in our system.}
\end{center}
\end{table}

\subsubsection{Adding a lexicon}
To add realistic lexicons to the text, we mined text summarization benchmark datasets such as CNN/Daily Mail \cite{DBLP:journals/corr/SeeLM17} as described in Section \ref{analyzer} and randomly chose a word based on its frequency.

\subsubsection{Adding a grammar}
To define grammars that generate practical text, we explored sentences appeared in text summarization benchmark datasets and created context free grammars accordingly. Then, we synthesized them to dynamically generate realistic document.

\subsection{Analyzer}
\label{analyzer}
For analyzer, we used a pretrained english model from Spacy\cite{spacy2} to extract named entities and (\textit{noun, noun\_modifier}) and (\textit{(verb, verb\_modifier}) relations and counted the number of occurrences of those words or relations.

\subsection{Generator}
\begin{figure}[htp]

\centering
\subfigure[An example of predefined grammar]{%
  \label{subfig:predefined_grammar}
  \includegraphics[clip=False, width=0.5\columnwidth]{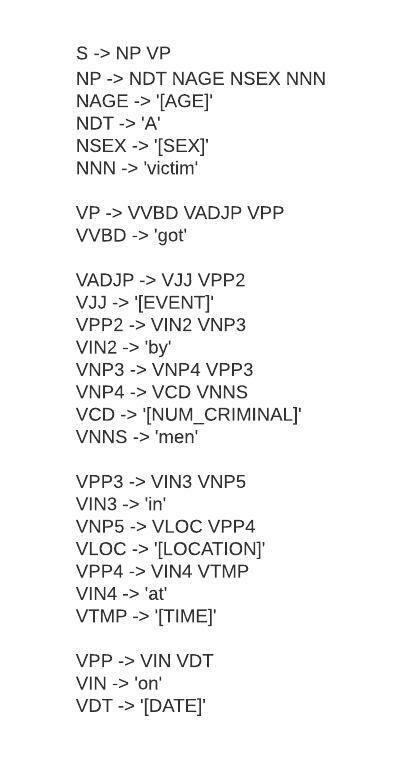}%
}

\subfigure[A visualization of predefined fact tree]{%
  \label{subfig:predefined_fact_tree}
  \includegraphics[clip=False,width=\columnwidth]{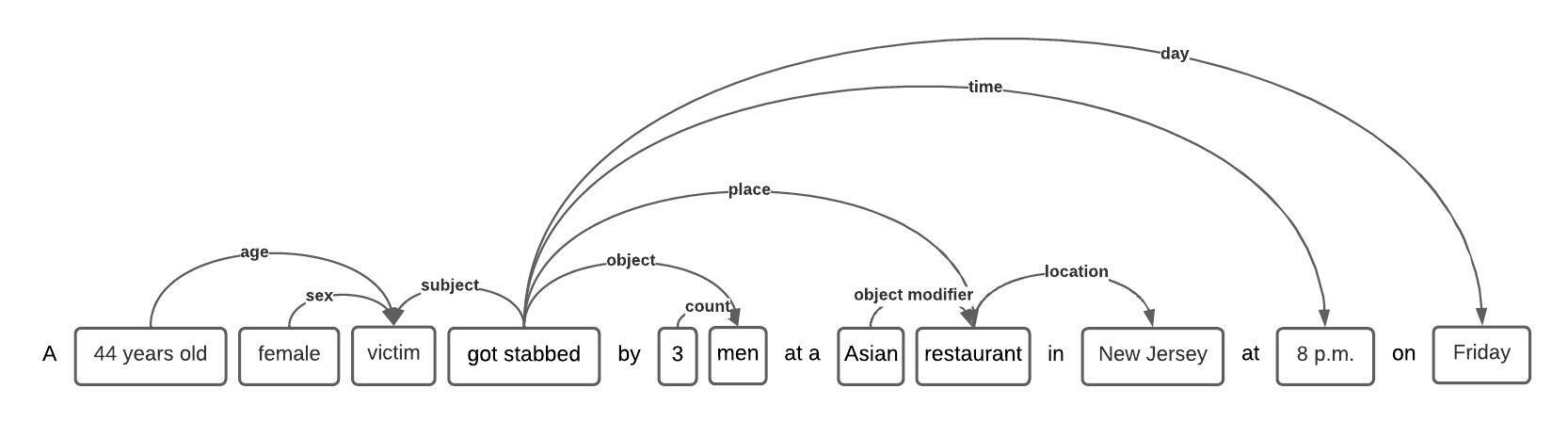}%
}

\caption{An example of a predefined grammar and corresponding abstract fact tree for the sentence \textit{"A 44 years old female victim got stabbed by 3 men at a Asian restaurant in New Jersey at 8 p.m. on Friday."}}
\label{fig:predefined_ex}
\end{figure}

\begin{figure}[htp]
  \centering
  \includegraphics[width=\linewidth]{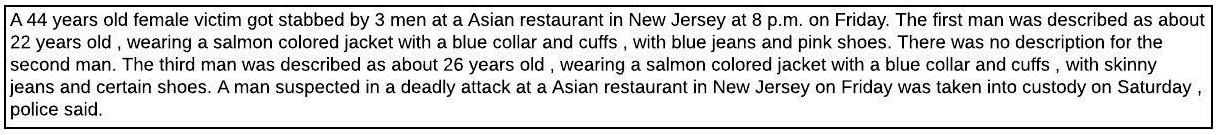}
  \caption{An example of Generated Text}
  \label{fig:generated_text}
\end{figure}

To generate a synthetic source document, we first manually predefine context free grammars and abstract fact trees as in \autoref{fig:predefined_ex}. In Figure \autoref{subfig:predefined_fact_tree}, the nodes of the tree are described as boxes and their attributes are marked with arrows. With the realistic facts from the analyzer and the abstract fact trees, we create a table of the facts that should appear in the synthetic article and fill the grammars with these facts. The table of the facts consists of an object (either \textit{noun} or \textit{verb}) and its attributes (\textit{modifiers}). For example, as in Figure \autoref{subfig:predefined_fact_tree}, an object, \textit{victim} has \textit{age}, \textit{sex}, and \textit{subject} as its attributes. Then, we parse the grammars and generate the documents. \autoref{fig:generated_text} exhibits the text generated from the generator.

\subsection{Summarizer}
\begin{figure}[htp]

\subfigure[System summary from PEGASUS-cnn-dailymail]{%
  \includegraphics[width=\linewidth]{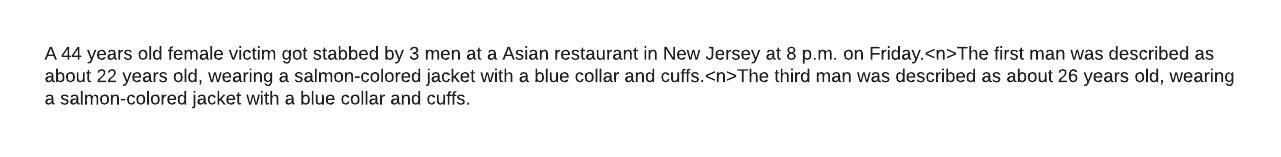}%
}
\subfigure[System summary from PEGASUS-xsum]{%
  \includegraphics[width=\linewidth]{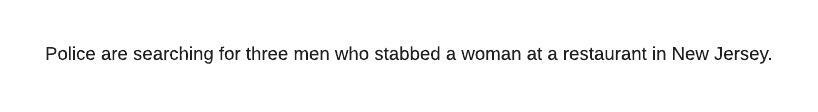}%
}
\subfigure[System summary from BART-large-cnn]{%
  \includegraphics[width=\linewidth]{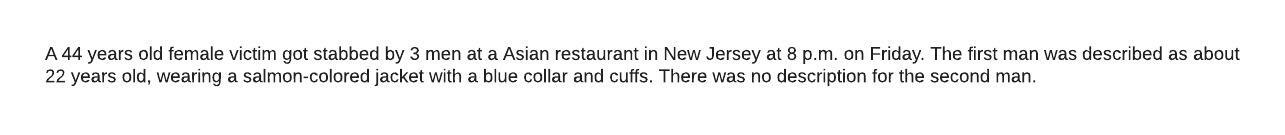}%
}
\subfigure[System summary from BART-large-xum]{%
  \includegraphics[width=\linewidth]{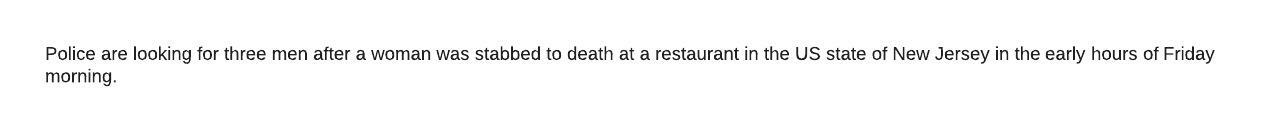}%
}
\caption{Output summaries of the synthetic document from the deep summarization models}
\label{fig:output_summaries}
\end{figure}

To summarize the synthetic source document, we utilize four state of the art deep text summarization models: BART-large-cnn, BART-large-xsum\cite{DBLP:journals/corr/abs-1910-13461}, PEGASUS-cnn-dailymail, and PEGASUS-xsum\cite{zhang2019pegasus}. However, our system is not limited to only these text summarization models but can be expanded to any text summarization models. \autoref{fig:output_summaries} shows the summary of the generated document with these deep summarizers. 

\subsection{Fact Extractor}
With a system generated summary, we utilize dependency parsing from Spacy\cite{spacy2} and extract facts from the summary based on the same factual relations used to represent facts in the text.


\subsection{Evaluator}
Given the table of facts in the source document and that of those in the summary, we measured the factual consistency, comprehensiveness, and compression rate. To find the overlaps between a source document and its abstractive summary, we checked if two nouns or verbs are synonyms or antonyms or how similar their senses are using wordnet. Also, since wordnet doesn't support the similarity measurement between adjectives or adverbs, we used glove word embeddings to measure their relatedness.

Factual consistency is the fraction of the facts in the summary that are correct. Correctness means that they also appear in the source. The facts in the summary contain both intrinsic and extrinsic factually inconsistent facts.
\begin{equation}
Factual\ Consistency = \frac{|\textrm{source-summary\ facts\ overlap}|}{|\textrm{facts\ in\ summary}|}
\end{equation}
and, comprehensiveness is the fraction of the facts in the summary that also appear in the source document.
\begin{equation}
Comprehensiveness = \frac{|\textrm{source-summary\ facts\ overlap}|}{|\textrm{facts\ in\ source}|}
\end{equation}
Compression rate is defined as introduced in \cite{napoles-etal-2011-evaluating}:
\begin{equation}
Compreesion\ rate = \frac{|\textrm{tokens\ in\ summary}|}{|\textrm{tokens\ in\ source}|} \times 100
\end{equation}

\section{Discussion}
In this section, we discuss the strengths and limitations of our system compared to existing evaluation systems/metrics and the reasoning behind choosing three different evaluation criteria: factual consistency, comprehensiveness, and compression rate in evaluating the text summarization models. 

Since we create synthetic documents, we have a natural interpretability and control in understanding and extracting facts in the documents. For example, in these two sentences: \textit{"A woman is 44 years old. A woman is killed"}, it is extremely difficult to predict that the woman is the same entity with the deep learning as they are ambiguous; however, our system can handle this issue as we generate the text based on the facts of \textit{(woman, 44 years old) and (woman, killed)}. Another strength of our system is that it considers the partial facts. For example, in a given sentence of \textit{"A man is described wearing a blue jacket and black jeans."}, our system both considers two partial facts: \textit{"A man is wearing a blue jacket."} and \textit{"A man is wearing black jeans."} and measure its quality accordingly.

However, our system is limited in matching implicit facts that appear differently in the source document and its summary. For example, considering these \textit{(subject, verb)} facts in \textit{"A woman is stabbed"} and \textit{"A woman is injured"}, even though \textit{(woman, injured)} implies \textit{(woman, injured)} to be true, it is extremely difficult to systematically describe that they are the same facts.

Our system considers three different criteria: factual consistency, comprehensiveness, and compression rate in evaluating the text summarization models because they are the crucial standards that a good summary should have. The goal of summarization is to produce a shorter text that captures the overall meaning of the source document. In this context, if the summary is not factually consistent or comprehensive, then its quality should be measured poorly. Also, even though the summary is comprehensive, it should be scored less if it is not much shorter than the source document – if the summary is exactly same as the source document, it should be penalized. Therefore, we utilized these three criteria to measure the quality of the system summaries generated from deep text summarization models.

Overall, our system describes the promising results as a new text evaluation metric system.
\section{Acknowledgements}
The authors thank professors Franz Kurfess, Alex Dekhtyar, and Stuart Russell for their valuable thoughts on this project.

\bibliographystyle{plain}
\bibliography{reference.bib}
\end{document}